\documentclass[runningheads]{llncs}

\usepackage[hyphens]{url}
\usepackage[hidelinks]{hyperref}
\usepackage[utf8]{inputenc}
\usepackage{subcaption}
\usepackage{mathabx}
\usepackage{microtype}
\usepackage{multirow}
\usepackage{blindtext}
\usepackage{graphicx}
\usepackage{cleveref}

\usepackage{layouts}

\crefformat{footnote}{#2\footnotemark[#1]#3}

\begin{document}

\title{Interactive Concept Mining on Personal Data -- Bootstrapping Semantic Services}
\titlerunning{Interactive Concept Mining on Personal Data}

\author{
	Markus Schröder \and
	Christian Jilek \and
	Andreas Dengel
}
\institute{
	Smart Data \& Knowledge Services Dept., DFKI GmbH, Kaiserslautern, Germany\\ \and
	Computer Science Dept., TU Kaiserslautern, Germany\\
	\email{\{markus.schroeder, christian.jilek, andreas.dengel\}@dfki.de}
}

\maketitle

\begin{abstract}
Semantic services (e.g. Semantic Desktops) are still afflicted by a cold start problem:
in the beginning, the user's personal information sphere, i.e. files, mails, bookmarks, etc., is not represented by the system.
Information extraction tools used to kick-start the system typically create 1:1 representations of the different information items.
Higher level concepts, for example found in file names, mail subjects or in the content body of these items, are not extracted.
Leaving these concepts out may lead to underperformance, having to many of them (e.g. by making every found term a concept) will clutter the arising knowledge graph with non-helpful relations.
In this paper, we present an interactive concept mining approach proposing concept candidates gathered by exploiting given schemata of usual personal information management applications and analysing the personal information sphere using various metrics.
To heed the subjective view of the user, a graphical user interface allows to easily rank and give feedback on proposed concept candidates, thus keeping only those actually considered relevant.
A prototypical implementation demonstrates major steps of our approach.

 \end{abstract}

\section{Introduction}
\label{sec:introduction}
Since the early 2000s, there has been a lot of research on Semantic Desktop (SD) \cite{SauermannBernardiDengel2005} systems.
A survey covering several relevant approaches can be found in \cite{DraganDecker2012}.
Especially in the last decade, ideas like linking such systems to the internet \cite{DraganDelbruGroza2011} or the (heterogeneous) IT environment of companies \cite{MausSchwarzDengel2013} have been pursued.
More recent approaches (e.g. DFKI's \textit{CoMem}\footnote{\url{https://comem.ai/}}) use such knowledge graphs to provide AI-powered support services like an SD-based work environment that re-organizes itself according to user contexts \cite{JilekSchroederSchwarz+2018}.
However, all these systems are still afflicted by a cold start problem:
in the beginning, the user's information sphere, i.e. files, mails, bookmarks, contacts, calendar entries, tasks, etc., is not represented by the system.
In the past, tools like Aperture\footnote{\url{http://aperture.sourceforge.net/}}  or Apache Tika\footnote{\url{https://tika.apache.org/}} have been used to extract the content of various information items and kick-start the system.
However, this typically led to a 1:1 representation: higher level concepts, e.g. found in file names, mail subjects or in the (content) body of these items, were not extracted.
On the one hand, we cannot spare these concepts, since SD systems may underperform without such additional vocabulary (for example experienced in \cite{JilekMausSchwarz+2015}).
On the other hand, accepting concepts that are not relevant for the user will clutter the created knowledge graph with irrelevant instances and non-helpful  relations between items.
Identifying the ``right'' concepts cannot be fully automated, since each user has a subjective view on their data \cite{Dengel2006}, deciding for themselves which concepts are actually relevant and which are not.
That is why we intend to follow an interactive (human-in-the-loop) approach for our concept extraction.

\section{Approach \& Demo}
\label{sec:approach}
\begin{figure}[t]
	\centering
	\includegraphics[width=1.0\linewidth]{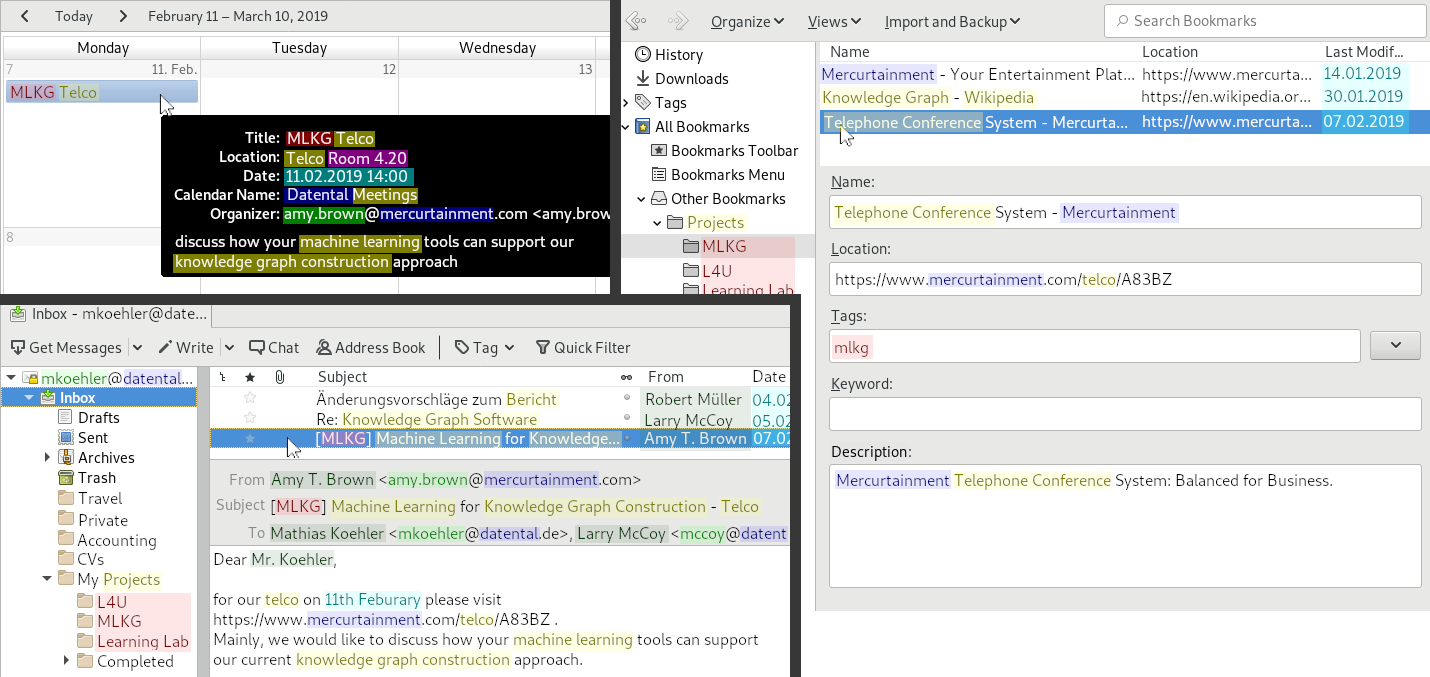}
	\caption{Excerpt of a personal information sphere: calendar, bookmarks and emails.}
	\label{fig:intro}
\end{figure}
In order to give a motivating scenario, Figure \ref{fig:intro} depicts a desktop system with a focus on three usual personal data silos: calendar (left), bookmarks (right) and mails (bottom):
The machine learning expert Mr. Koehler works together with Ms. Brown in the MLKG project, where he got an invitation for a telephone conference via mail.
While he uses his calendar to store meetings, the browser records his bookmarks. 
For a better understanding, we colour-highlight the following concepts occurring in his information sphere:
projects (red), persons (green), organizations (blue), times (cyan), places (purple) and general topics (yellow).
To ultimately form a personal information model (PIMO) \cite{SauermannVanElstDengel2007} and bootstrap semantic services, our approach semi-automatically extracts such concepts.
The main idea is to exploit the given schemata of personal information management  (PIM) applications, like those shown in Figure \ref{fig:intro}.
In the following we describe several ideas how to realize this vision.

Extracting information from rather structured text fields (e.g. found in the mail schema) benefits the enrichment of rather unstructured texts (e.g. folder names).
This is based on the assumption that certain text fields contain particular named entities more likely and with a similar format, making them easier to identify.
For instance, our mail application's ``From'' column contains first and last names, while the calendar's location field stores place-related information.
In domains of email addresses and links, organization names can be found more likely.
Once we identify a concept like the organization ``Mercurtainment'', gazetteer-based methods \cite{JilekSchroederNovik+2019} are able to discover similar occurrences, e.g. in the bookmark's title.

Typical terminology extraction approaches work on collections of unclassified text documents.
However, we can take advantage of terms appearing in certain PIM items.
While bookmark titles, for example, are often given by the website's author, titles for calendar events, emails (subject), files or folders are usually given by the user, especially providing a kind of manual summary.
Thus, they will more likely contain the user's own vocabulary, for instance, technical terms or abbreviations (acronyms).
Additionally, if such terms occur in different data silos independently (indicating that the user has interacted with the topic in various ways), we can conclude that they are more relevant.
``MLKG'', in our example an acronym for Mr. Koehler's current project, can be found multiple times in all data silos. This relevancy estimation is also transferable to spatial-based observations in PIM hierarchies like file, mail and bookmark folder trees.
Terms located higher in such hierarchies are often of a more general nature, since they early separate a hierarchical data space.
In our scenario this is the case for the term ``Projects'', which is located close to the root in the mail and bookmark folder trees.
Temporal relationships are also possible indicators for semantic relations across data silos.
In our example, ``11th Feburary'' is mentioned in a mail actually referring to an appointment taking place on that day (see calendar).
Additional implicit relations can be discovered by detecting similarly labelled (or copied) text content.
As an example, the hyperlink in our mail got bookmarked while a containing text snippet was copied to a calendar entry's description.
Apart from typical named entities, such as dates, person names, locations, etc., there are also PIM concepts \cite{SauermannVanElstDengel2007} which are best classified as general topics.
Once they are identified, they further help to classify the information item's content on a conceptual level.
In our scenario, topics like ``Machine Learning'', ``Telco'' and ``Knowledge Graph Construction'' provide a good abstraction of Mr. Koehler's current context.

To enable users to lift terms to relevant and typed concepts, they are supported with an interface presenting all extracted terms. However, the high amount of terms found in common information spheres triggers a lot of user effort to scan through them.
Thus, we suggest an appropriate filtering and ranking of terms for a more convenient selection.
Each term potentially has very different characteristics in various metrics.
Relevant terms come in various shapes and facets like (upper case) acronyms of project names, lower case last names in mail addresses, short infrequent folder names, long frequent multi-word terms in mail bodies, etc.
Hence, sorting concept candidates by just one metric will not be sufficient.
In order to enable different perspectives, a weighted harmonic mean score using appropriate combinations of multiple metrics with individual weights can be calculated. However, users will likely feel overwhelmed by the high variety of settings.
To counter this, the selection process can be further eased by providing
widely proven presets discovered through user studies.

\noindent
\textbf{Demo.}
\begin{figure}[t]
	\centering
	\includegraphics[width=1.0\linewidth]{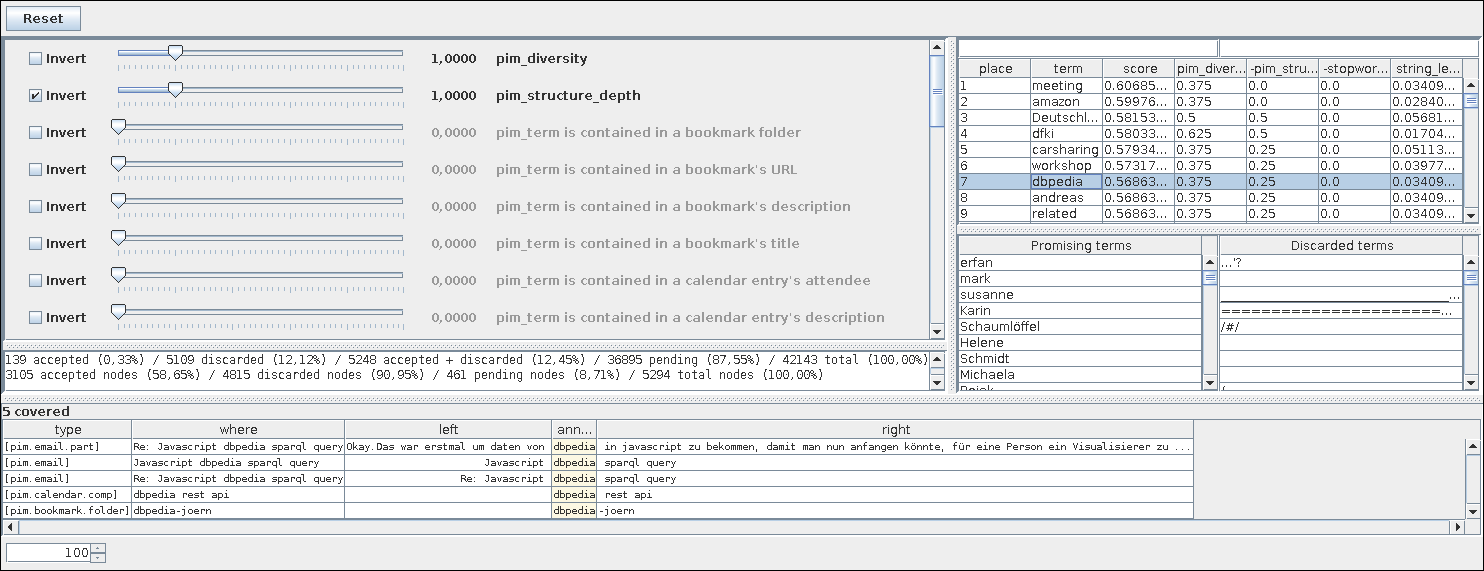}
	\caption{
Prototypical implementation of our concept mining tool.
	}
	\label{fig:gui}
\end{figure}
Major steps of our approach are already implemented in our demo application.
After a PIM dataset is crawled, our tool preprocesses each data silo:
persons are automatically identified using names found in email addresses. 
Mail bodies are converted to plain text if necessary (attachments are not considered, yet).
Cities and countries are discovered in calendar entries' location field using a gazetteer-based approach.
All texts in email folders, subjects, bodies, calendar summaries, descriptions, locations as well as bookmark folders, titles and descriptions are tokenized to gather (single-word) terms.

Subsequently, the collected terms are presented to the user in a graphical user interface (see Figure \ref{fig:gui}).
In advance, terms containing solely symbols or numbers are automatically discarded, while found person names, cities and countries are marked as promising.
The freely configurable combinations (left side) enable the user to rank terms individually.
After a reasonable combination is found, users can binary classify sorted terms (right side) as either ``discarded'' (by pressing \textit{delete}) or ``promising'' (by pressing \textit{enter}). A text field (middle) indicates the current progress: besides counts about classified terms, the interface also calculates the number of (not) covered mails, bookmarks and calendar entries by terms marked as promising.
Terms can be selected to view their occurrence in the whole data set (bottom).
A more precise presentation could be retrieved with deep links \cite{SchroederJilekDengel2018}.

On our demo page\footnote{\url{https://www.dfki.uni-kl.de/~mschroeder/demo/pim-semantifier/}}, the system's prototype together with a personal information crawler can be downloaded.
Additionally, a demo video presents how the tools are used.

\section{Conclusion \& Outlook}
\label{sec:conclusion}
Motivated by the cold start problem of Semantic Desktop systems, this paper outlined an approach for extracting higher level concepts from personal information spheres to bootstrap semantic services.
By means of an exemplary scenario, we proposed several ideas on how to exploit given schemata of usual personal information management applications such as mail, calendar and bookmark clients.
Our corresponding user interface concept is designed to allow a user to conveniently rank and select potential concepts from several extracted candidate terms.
A first prototypical implementation is presented in a demo.

In the future, we will additionally consider files, another big and often unstructured part of the personal information sphere.
Based on the extracted (and verified) concepts found with our tool, we also plan to perform user context mining, especially taking interconnections of the different data silos into account.
Both aspects exceed this paper's scope and will be addressed in dedicated papers.
\\
\\
\noindent
\textbf{Acknowledgements.}
Parts of this work have been funded by
the German Federal Ministry of Food and Agriculture in the \textit{SDSD} project [2815708615]
and by
the Deutsche Forschungsgemeinschaft (DFG) in the \textit{Managed Forgetting} project [DE 420/19-1].
We also thank
Heiko Maus,
Sven Schwarz,
Jörn Hees,
and
Ansgar Bernardi
for their feedback and the fruitful discussions. 
\bibliographystyle{llncs} 	
\bibliography{paper}

\end{document}